\documentclass{article}
\PassOptionsToPackage{numbers, compress}{natbib}
\usepackage[preprint]{neurips_2020}
\usepackage[utf8]{inputenc} 
\usepackage[T1]{fontenc}    
\usepackage{hyperref}       
\usepackage{url}            
\usepackage{booktabs}       
\usepackage{amsfonts}       
\usepackage{nicefrac}       
\usepackage{microtype}      
\usepackage{graphicx}
\usepackage{array}
\title{Zoom, Enhance! Measuring Surveillance GAN Up-sampling}

\author{
  Abdalla Al-Ayrot\\
  Stanford University\\
  \texttt{abdalla@stanford.edu} \\
  \And
  Jake Sparkman\\
  Stanford University\\
  \texttt{jdspark@stanford.edu} \\
  \AND
  Utkarsh Contractor\\
  Stanford University\\
  \texttt{utkarshc@stanford.edu} \\
}

\begin{document}

\maketitle
\begin{abstract}
Deep Neural Networks\cite{DBLP:journals/corr/Schmidhuber14} have been very successfully used for many computer vision and pattern recognition applications. While Convolutional Neural Networks (CNNs)\cite{DBLP:journals/corr/Schmidhuber14} have shown the path to state of art image classifications, Generative Adversarial Networks or GANs\cite{goodfellow2014generative} have provided state of art capabilities in image generation. In this paper we extend the applications of CNNs and GANs to experiment with up-sampling techniques in the domains of security and surveillance. Through this work we evaluate, compare and contrast the state of art techniques in both CNN and GAN based image and video up-sampling in the surveillance domain. As a result of this study we also provide experimental evidence to establish DISTS\cite{DBLP:journals/corr/abs-2004-07728} as a stronger IQA\cite{IQAcomp} metric for comparing GAN Based Image Up-sampling in the surveillance domain.
\end{abstract}

\section {Introduction}
There has been much work done on video and image super resolution, where previous approaches have examined up-sampling of both images and videos. However, there is limited literature on up-sampling images and videos in the surveillance domains, especially up-sampling low resolution security footage. Furthermore most publications highlight SSIM\cite{SSIM}\cite{Ding_2021} as one of the dominant metric for Image Quality Assessment (IQA)\cite{IQAcomp}\cite{DBLP:journals/corr/ShiCHTABRW16}\cite{DBLP:journals/corr/LedigTHCATTWS16}\cite{Ding_2021} of upsampled images. 

We present a study and evaluation in using Generative Adversarial Networks (GANs) to upscale low resolution surveillance videos. Through this endeavour we put forward two important contributions:
\begin{itemize}
    \item Establishing DISTS as a stronger metric to evaluate the up-sampling quality of surveillance videos; we show DISTS to be superior to both SSIM\cite{SSIM} and LPIPS\cite{DBLP:journals/corr/abs-1801-03924} when trying to predict MOS\cite{MOS} scores.
    \item Contributing a study and evaluation of state of the art GAN based image and video up-sampling in the surveillance domain; and subsequently contributing a up-sampled dataset of surveillance videos using the state of the art SRGAN\cite{DBLP:journals/corr/LedigTHCATTWS16} architecture. 
\end{itemize}

We hope the contributions in this paper provides a good baseline for evaluation and measurements for surveillance video up-sampling and can be used to further research in the field.

\section{Dataset}
There exist a number of low resolution video anomaly datasets
used by the deep learning community for research. However few surveillance video datasets exist with HR videos.  Our focus in this study will be on up-scaling surveillance videos from the UCF Video Anomaly Detection Dataset\cite{DBLP:journals/corr/abs-1801-04264} and PETS 2007\cite{pets_2007} so we can provide the research community with the evaluation on two industry accepted surveillance datasets and also a high quality video anomaly dataset. We use the Video Anomaly Detection\cite{DBLP:journals/corr/abs-1801-04264} and PETS 2007\cite{pets_2007} dataset for training, validation and testing. The Video Anomaly Detection\cite{DBLP:journals/corr/abs-1801-04264} dataset consists of 128 hours of videos consisting of 1900 real-world surveillance videos, with 13 realistic anomalies such as fighting, road accident, burglary, robbery, etc. as well as normal activities captured by surveillance cameras. The videos are long untrimmed surveillance videos with low resolution with large intra-class variations due to changes in camera viewpoint and illumination, and background noise, thus making it a great fit for our use case. For our experiments we use a miniature version of this dataset with 2207 images, split into 1200 training, 507 validation images and 500 images in the test set. Few samples from the dataset can be seen in Figure 1. Additionally we also use the PETS 2007\cite{pets_2007} dataset which is a collection of images from four surveillance cameras with varied perspective to provide examples scenes such as general loitering, theft, abandoned luggage, etc.

\begin{figure}
\includegraphics[scale=0.5]{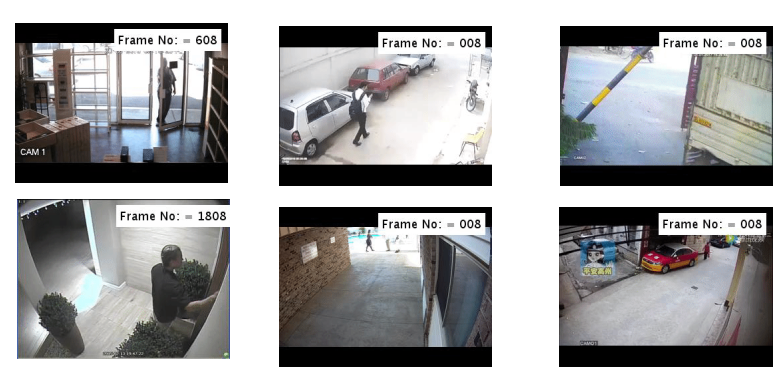}
\caption{Sample frames from Video Anomaly Detection Dataset \protect\cite{DBLP:journals/corr/abs-1801-04264}}
\end{figure}

\section{Approach and Methodology}
\subsection{Choice of Neural Architecture}

Super-resolution (SR) is used to upscale an image or video from a low-resolution (LR) image or video to a high-resolution (HR) image or video. The method is essentially estimating the HR image or video from its LR equivalent. 

Advances in the research of learning-based methods, have increased the performance of single-image SR (SISR) significantly. Applying the advances in SISR to video SR (VSR) however is more challenging, as simply taking SISR over each video frame, can lead to inferior results due to temporal coherency issues. Recent research has shown that combining state-of-the-art SISR and (multi image SR) MISR methods with a spatio-temporal approach surpasses state-of-the-art VSR results with the introduction of iSeeBetter.\cite{DBLP:journals/corr/abs-2006-11161}

To this effect, we have chosen to experiment with the following approaches:
\begin{itemize}
    \item Baseline Approach: Sub-pixel CNN Based Video Up-sampling
    \item SOTA Approach 1: Enhanced SRGAN Based Video Up-sampling
    \item SOTA Approach 2: Recurrent Back-Projection Networks (RBPN) + SRGAN Based Video Up-sampling
\end{itemize}

\subsection{Baseline Approach - Sub-pixel CNN Based Architecture}
\subsubsection{Model Details and Architecture}
Our baseline approach follows work done by Shi et al. on Real-Time Single Image and Video Super-Resolution Using an Efficient Sub-Pixel Convolutional Neural Network \cite{DBLP:journals/corr/ShiCHTABRW16} which is the first convolutional neural network (CNN) capable of real-time SR of 1080p videos on a single K2 GPU. In this approach, contrary to previous works, the authors propose to increase the resolution from LR to HR only at the very end of the network and super-resolve HR data from LR feature maps. This eliminates the need to perform most of the SR operation in the far larger HR resolution. For this purpose, this approach proposes an efficient sub-pixel convolution layer to learn the up-scaling operation for image and video super-resolution. For the baseline version architecture we will be using l = 3, (f1, n1) = (5, 64), (f2, n2) = (3, 32) and f3 = 3 in our evaluations as proposed in the paper\cite{DBLP:journals/corr/ShiCHTABRW16}. The choice of the parameter is inspired by SRCNN’s \cite{DBLP:journals/corr/DongLHT15} 3 layer 9-5-5 model. 

\begin{figure}
\includegraphics[scale=0.3]{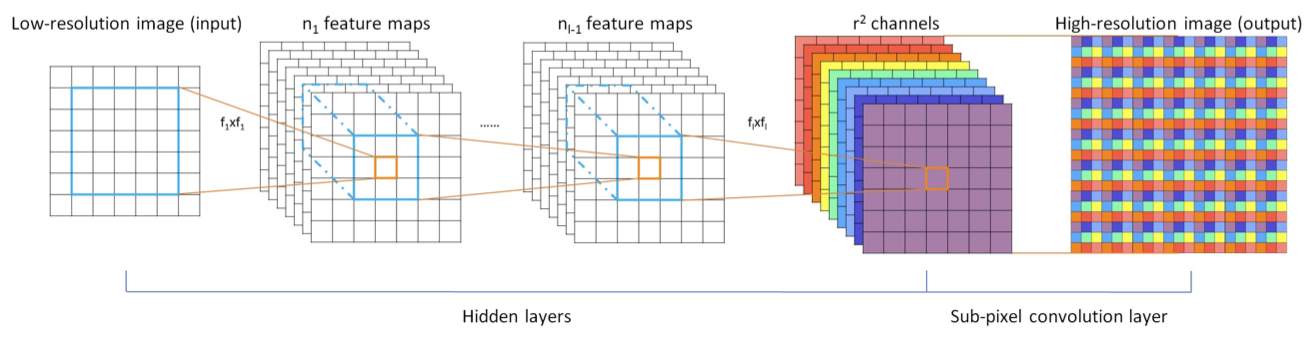}
\caption{The proposed efficient sub-pixel convolutional neural network (ESPCN), with two convolution layers for feature maps extraction, and a sub-pixel convolution layer that aggregates the feature maps from LR space and builds the SR image in a single step.}
\end{figure}

\subsubsection{Training and Scoring Details}

In the training phase, 17r × 17r pixel sub-images are extracted from the training ground truth images IHR, where r is the up-scaling factor. The sub-images are extracted from original images with a stride of \( (17 - \sum mod (f, 2)) \times r \) from \( I^{HR}\) and a stride of \(17 - \sum mod (f, 2) \) from \( I^{LR}\). This ensures that all pixels in the original image appear once and only once as the ground truth of the training data\cite{DBLP:journals/corr/ShiCHTABRW16}. The training stops after no improvement of the cost function is observed after 100 epochs. Initial learning rate is set to 0.01 and final learning rate is set to 0.0001 and updated gradually when the improvement of the cost function is smaller than a threshold µ. The training takes roughly about 12 hours on a 1080 Titan X Pascal GPU on images from UCF Mini Image Dataset \cite{sultani2019realworld} for upscaling factor of 4 and 100 epochs. The train and validation charts can been seen in Figure 3.

\begin{figure}
\includegraphics[scale=0.22]{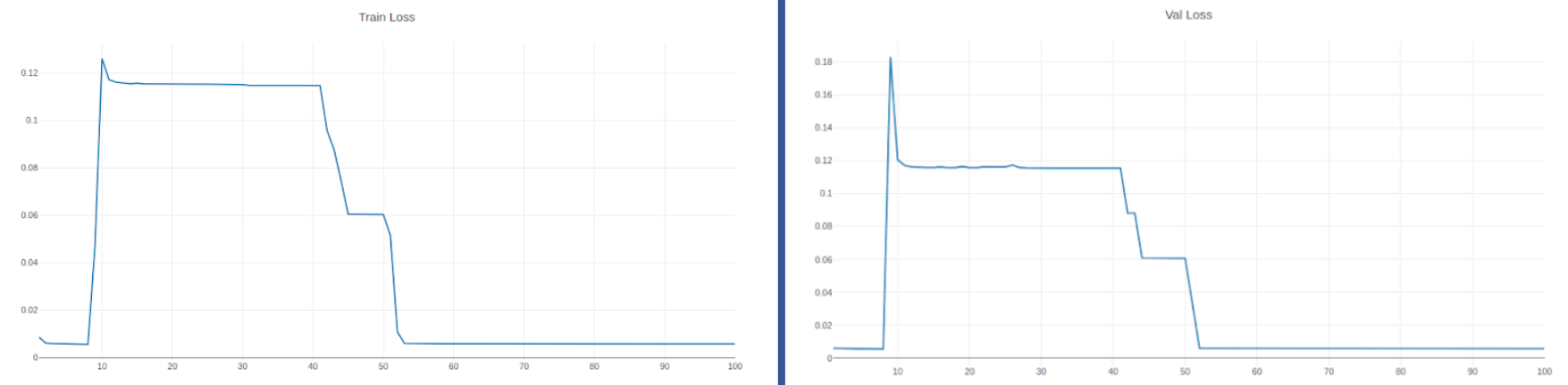}
\caption{Train and Validation loss for sub-pixel convolutional neural network (ESPCN).}
\end{figure}

\subsection{SOTA Approaches: GAN Based Up-Sampling}
\subsubsection{Enhanced SRGAN - Model Details and Architecture}
Our first proposed approach is inspired by Ledig et al. on Photo-Realistic Single Image Super-Resolution Using a Generative Adversarial Network \cite{DBLP:journals/corr/LedigTHCATTWS16}. This was first framework capable of inferring photo-realistic natural images for 4x up-scaling factors using a deep residual network to recover photo-realistic textures from heavily down-sampled images. In this approach a super-resolution generative
adversarial network (SRGAN) for which is applied a
deep residual network (ResNet) with skip-connection and
diverge from MSE as the sole optimization target. Different
from previous works, this approach defines a novel perceptual loss using high-level feature maps of the VGG network\cite{simonyan2015deep} combined with a discriminator that encourages solutions perceptually hard to distinguish from the HR reference images. 

At the core of this very deep generator network G, which is illustrated in Figure 3 are B residual blocks with identical layout. Inspired by Johnson et al. \cite{johnson2016perceptual} employed are the block
layout. Specifically, two convolutional layers with small 3×3 kernels and 64 feature maps followed by batch-normalization layers and ParametricReLU \cite{he2015delving} as the activation function. The resolution of the input image is increased with two trained
sub-pixel convolution layers as proposed by Shi et al. \cite{DBLP:journals/corr/ShiCHTABRW16}.
To discriminate real HR images from generated SR
samples a discriminator network is trained. The architecture
is shown in Figure 4. The architectural guidelines followed are summarized by Radford et al. \cite{radford2016unsupervised} and use LeakyReLU
activation (\begin{math}\alpha = 0.2\end{math}) and avoid max-pooling throughout
the network. The discriminator network contains eight
convolutional layers with an increasing number of 3 × 3
filter kernels, increasing by a factor of 2 from 64 to 512 kernels as in the VGG network \cite{simonyan2015deep}. Strided convolutions are used to reduce the image resolution each time the number of features is doubled. The resulting 512 feature maps are followed by two dense layers and a final sigmoid activation function to obtain a probability for sample classification.

\begin{figure}
\includegraphics[scale=0.4]{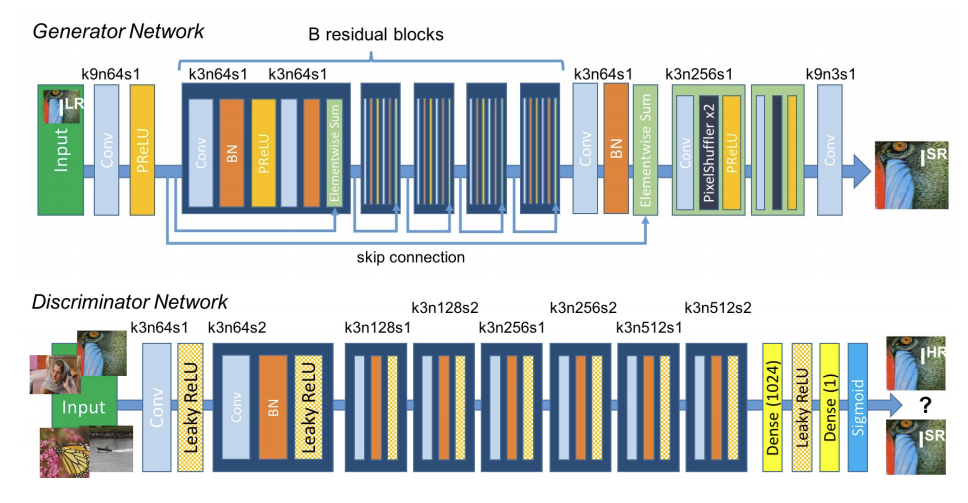}
\caption{ Architecture of Generator and Discriminator Network with corresponding kernel size (k), number of feature maps
(n) and stride (s) indicated for each convolutional layer.}
\end{figure}

\paragraph{Training and Scoring  Details}
For our v1 SRGAN version we have run the model on our UCF\cite{sultani2019realworld} mini dataset. All experiments are performed with a scaling factor of ×4 between LR and HR images. We obtain LR images by down-sampling HR images using the MATLAB bicubic kernel function. The mini-batch size is set to 16. The spatial size of cropped HR patch is 128 × 128. Training a deeper network benefits from a larger patch size, since an enlarged receptive field helps to capture more semantic information. However, it costs more training time and consumes more computing resources. The training process is divided into two stages. First, train a PSNR-oriented model with the L1 loss. , then employ the trained PSNR-oriented model as an initialization for the generator. The learning rate is set to 0.0001 and halved at [10, 30 , 50, 75] iterations. 

\begin{figure}
\begin{center}
\includegraphics[scale=0.3]{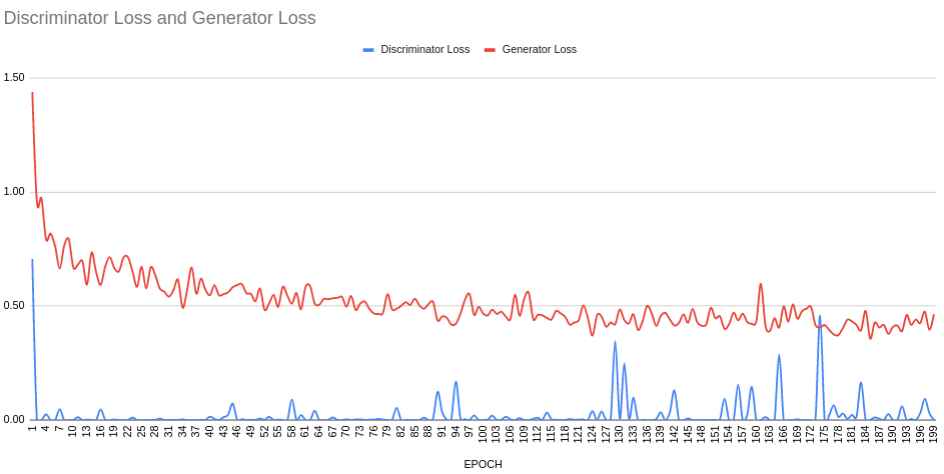}
\caption{GAN losses for Enhanced SRGAN training}
\end{center}
\end{figure}

Pre-training with pixel-wise loss helps GAN-based methods to obtain more visually pleasing results. The reasons are that 1) it can avoid undesired local optima for the generator; 2) after pre-training, the discriminator receives relatively good super-resolved images instead of extreme fake ones (black or noisy images) at the very beginning, which helps it to focus more on texture discrimination\cite{DBLP:journals/corr/LedigTHCATTWS16}. For optimization, we use Adam with \begin{math} \beta 1 = 0.9, \beta 2 = 0.999 \end{math}. We implement our models with the PyTorch framework and train them using NVIDIA Titan X Pascal GPUs. The SRGAN training is very computational expensive and takes about 36 hours on a Titan X Pascal 1080 GPU for 200 EPOCHS for 1707 images from UCF Mini dataset \cite{sultani2019realworld}. The Generator and Discriminator (GAN) losses are seen in Figure 5.

\subsubsection{iSeeBetter - Model Details and Architecture}
This approach used iSeeBetter \cite{DBLP:journals/corr/abs-2006-11161}. The framework achieved state-of-the-art results by combining Recurrent Back-Projection Networks (RBPN) as its generator and the discriminator from SRGAN. Figure 6 shows the original architecture of the iSeeBetter-network  (see Table 1 for adopted notation). The RBPN generator preserves spatio-temporal information by combining SISR and MISR. The horizontal flow of the network (illustrated by the blue lines in Figure 6) upsamples \textit{LR\textsubscript{t}} using SISR, with a DBPN architecture \cite{haris2018deep}. Up-down-up sampling is performed using 8 x 8 kernels with a stride of 4 and a padding of 2. As with enhanced SRGAN we use ParametricReLU \cite{he2015delving} as the activation function. The vertical flow of the network (illustrated by the red arrows in Figure 6) performs MISR by utilizing a ResNet Architecture. We use three tiles of five blocks each consisting of two convolutional layers with 3 × 3 kernels, padding of 1 and stride of 1. As with enhanced SRGAN and the DBPN architecture we use ParametricReLU \cite{he2015delving} as the activation function. The MISR computes the residual features from \textit{LR\textsubscript{t}}, its neighboring
frames (\textit{LR\textsubscript{t - 1}}, ..., \textit{LR\textsubscript{t - n}}) and the precomputed dense motion flow maps (\textit{F\textsubscript{t - 1}}, ..., \textit{F\textsubscript{t - n}}).

\begin{figure}
\includegraphics[scale=0.35]{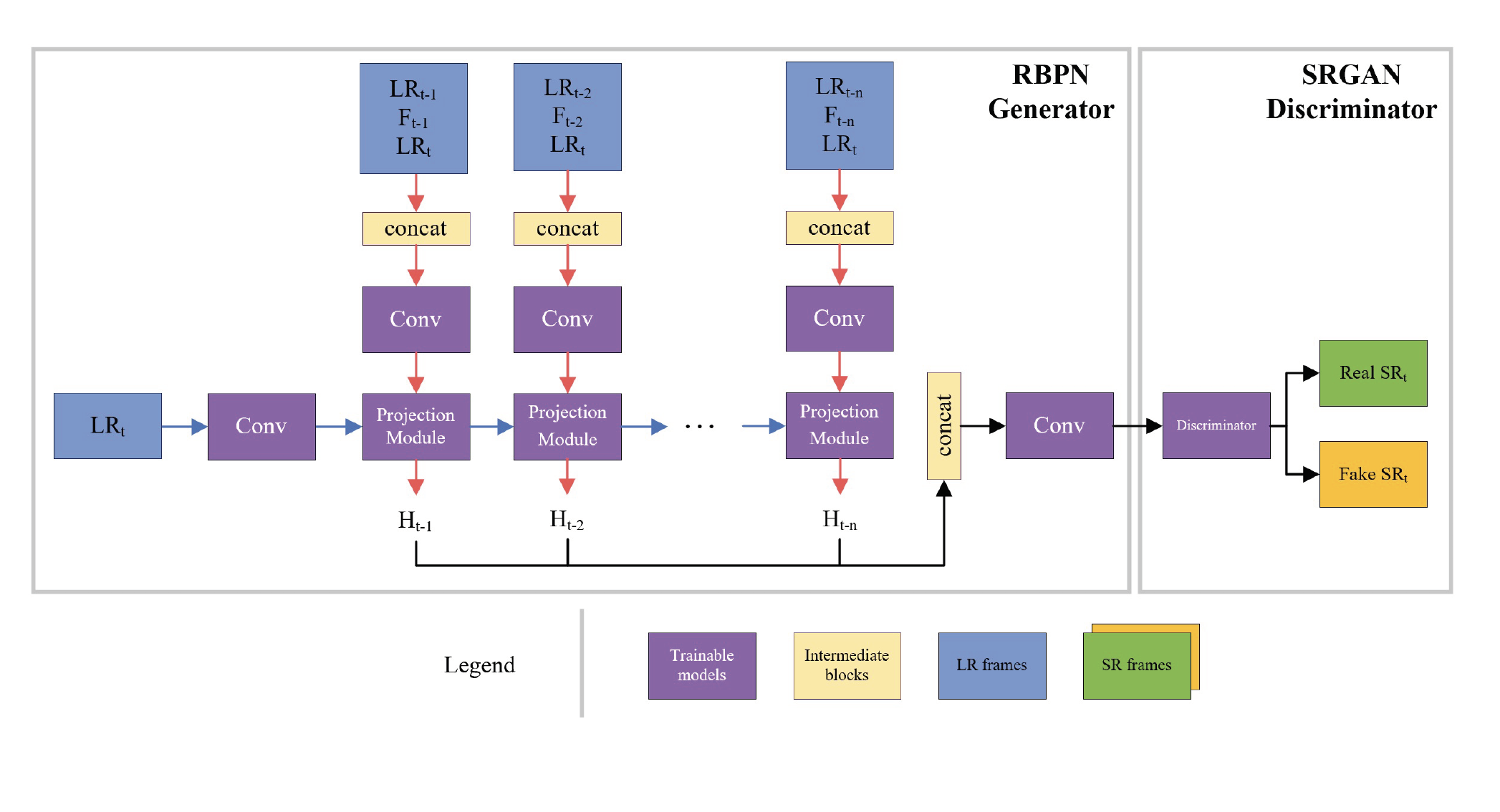}
\caption{Overview of iSeeBetter  \protect\cite{DBLP:journals/corr/abs-2006-11161}}
\end{figure}

\begin{table}
\centering
\small
\caption{Adopted notation for iSeeBetter
\protect\cite{DBLP:journals/corr/abs-2006-11161} }
\def\arraystretch{1.2}%
\begin{tabular}{m{1cm} m{5cm}}
 \textit{HR\textsubscript{t}} & input high resolution image \\ 
 \textit{LR\textsubscript{t}} & low resolution image (derived from \textit{HR\textsubscript{t}})\\ 
 \textit{F\textsubscript{t}} & optical flow output \\ 
 \textit{H\textsubscript{t - 1}} & residual features extracted from (\textit{LR\textsubscript{t - 1}}, \textit{F\textsubscript{t - 1}},  \textit{LR\textsubscript{t}}) \\ 
 \textit{SR\textsubscript{t}} & estimated HR output \\
\end{tabular}
\end{table}

RBPN detects missing  information from \textit{LR\textsubscript{t}} at each projection stage and recovers details by extracting residual features from neighboring frames. As a result, the convolutional layers that feed the projection modules in Figure 6 act as feature extractors.

\begin{figure}
\begin{center}
\includegraphics[scale=0.5]{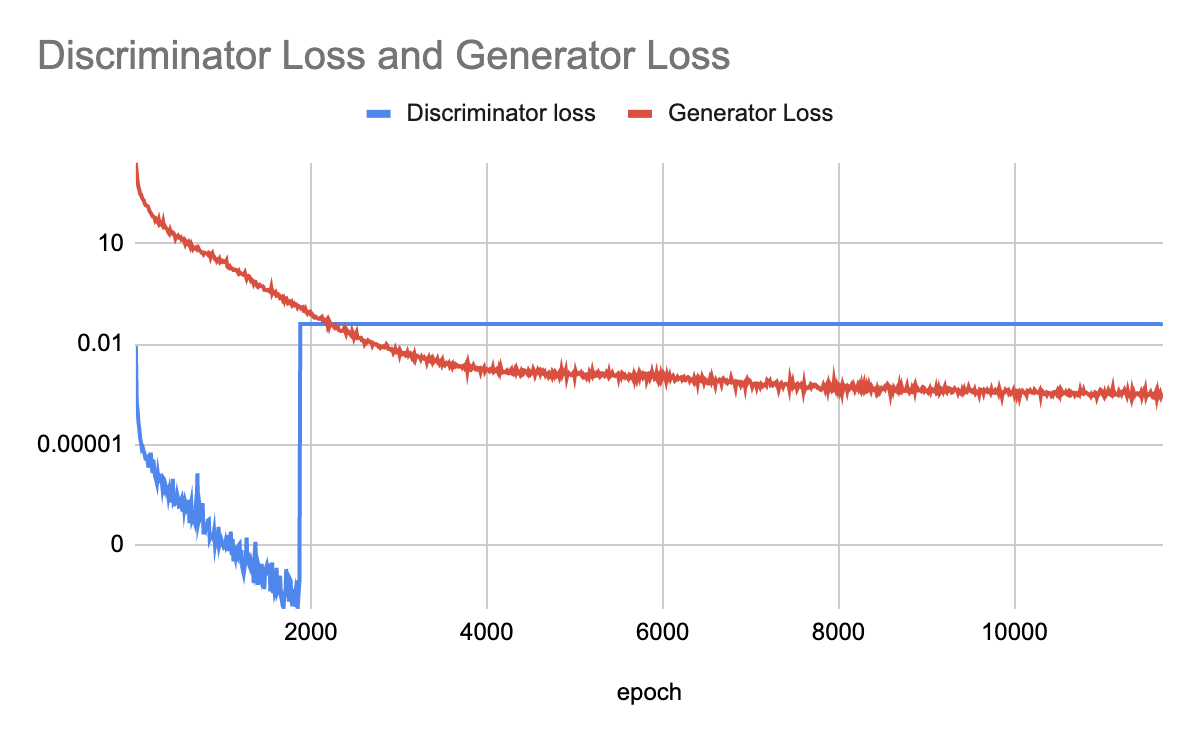}
\caption{GAN losses for iSeeBetter training}
\end{center}
\end{figure}

\paragraph{Training and Scoring  Details} We ran the model on our UCF \cite{sultani2019realworld} mini dataset for our v1 iSeeBetter edition. The LR and HR images are scaled by a factor of four in all experiments. We get LR images by using the MATLAB bicubic kernel function to down-sample HR images. The size of the mini-batch is set to one. The cropped HR patch has a spatial size of 32x32. A larger patch size helps to train a deeper network since a larger receptive area helps to collect more semantic knowledge. However, it takes longer to practice and uses more computational power.

MSE is the most widely used loss function in a wide range of state-of-the-art SR methods that seek to increase an image's PSNR to determine image quality. \cite{Ward_2017} Optimizing for MSE during training is widely known to increase PSNR and SSIM.
These metrics, however, may fail to capture fine details in the image, ensuing in a misrepresentation of perceptual quality. \cite{DBLP:journals/corr/LedigTHCATTWS16}
 The reason for this it was found in some experiments that some manually distorted images had an MSE score comparable to the original image.\cite{DBLP:journals/corr/abs-2006-11161}
We train the model with the four loss functions  originally proposed in iSeeBetter (MSE, perceptual,
adversarial, and TV) and weight the results for each frame. We use the PyTorch framework to build our models, and we train them on NVIDIA Tesla V100 GPUs. The UCF Mini dataset \cite{sultani2019realworld} was used to train the model.  The Generator and Discriminator (GAN) losses are seen in Figure 6.

\section{Evaluation Methodology/Results}
\begin{figure}
\begin{center}
\includegraphics[scale=0.3]{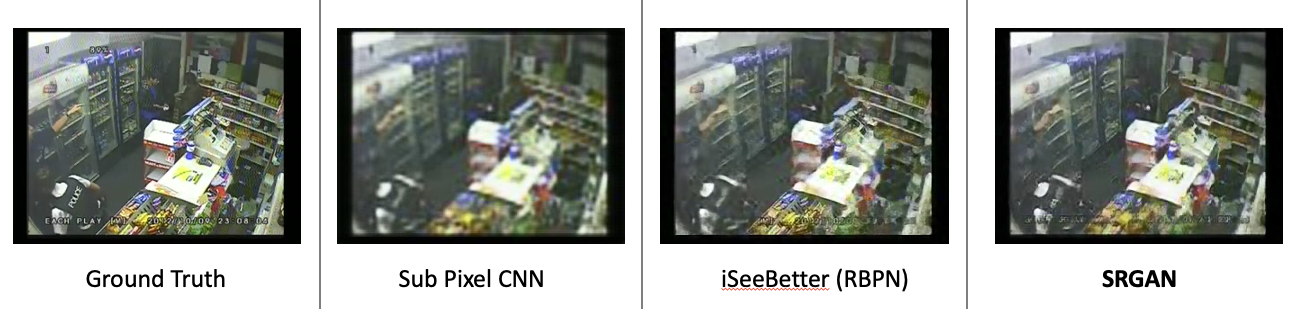}
\caption{Comparison of our model performance across various neural architectures.}
\end{center}
\end{figure}

Objective image quality assessment is still a challenging problem, and due to this we decided to use Mean Opinion Scores from humans to serve as the ground truth for image quality assessment for the various up-sampling methods. However, using MOS to quantify image quality is expensive and time consuming. To help reduce our reliance on MOS, we conducted a literature review to understand what other authors were using for objective image quality assessment. During this review, we found that many authors are using sub-optimal metrics for prediction of MOS. The metrics we evaluated are detailed below:

\begin{enumerate}
\item MOS, or Mean Opinion Score, is the average of human judgements on the quality of the image. This metric is based entirely on human perception, and as such it is dependent on a variety of environmental and idiosyncratic factors specific to the person that affect the perception of the image. 
\item SSIM, or Structural Similarity (SSIM) index \cite{SSIM}, has become a standard way of measuring IQA for a variety of applications. SSIM was motivated by an understanding of the image quality assessment problem as being more complicated than mirroring a variety of known properties of the human visual system. It seeks to better measure quality through an addition of the assessment of the degradation of structural information. This is motivated by an understanding that the human visual system is well adapted to extract structural information from an image. It compares luminance, contrast, and structural similarity between two images as these three components are largely independent.  
 \item LPIPs, or Learned Perceptual Image Patch Similarity \cite{zhang2018perceptual} was first proposed after the authors realized that deep networks - even across architectures and supervision type – provides an embedding that agrees surprisingly well with human perception. They then "calibrated" feature responses from the internal activations of various networks trained for high-level image classification tasks. Specifically, the LPIPS metric extracts features from L layers, which are then unit normalized in the channel dimension. These are then scaled channel wise, and the \(l_{2}\) distance is computed. Finally, these are averaged spatially and summed channel wise to generate an image specific distance metric. To compare images, one image's distance metric is compared to another via a small network that is trained to predict perceptual judgment. We used the open source implementation provided by \cite{zhang2018perceptual} and have chosen to report scores as \(1 - LPIPS_{dist}\).
\item DISTS, or Deep Image Structure and Texture Similarity metric  \cite{DBLP:journals/corr/abs-2004-07728}, is also based on deep network responses but is explicitly designed to tolerate texture re-sampling (e.g., replacing one patch of grass with another). DISTS is based on an injective mapping function built from a variant of the VGG network, and combines SSIM-like structure (using global correlations) and texture (using global means) similarity measurements between corresponding feature maps of the two images. It is sensitive to structural distortions but at the same time robust to texture re-sampling and modest geometric transformations. The fact that it is robust to texture variance is also helpful when evaluating images generated by GANs. We again used an open source implementation provided by Ding et al\cite{IQAcomp}. As DISTS maps images to a perceptual distance space, a score closer to 0 corresponds to more similar images. We have chosen to report scores as \(1 - DISTS_{dist}\) to allow for an intuitive comparison.
\item Fréchet Inception Distance(FID)\cite{FID_NIPS2017_8a1d6947} which captures the similarity of generated images to real ones. FID is consistent with increasing disturbances and human judgment and it captures the similarity of generated images to real ones better than the Inception Score\cite{DBLP:journals/corr/SalimansGZCRC16}\cite{FID_NIPS2017_8a1d6947}.
\end{enumerate}

\begin{figure}[!htbp]
\begin{center}
\includegraphics[scale=0.35]{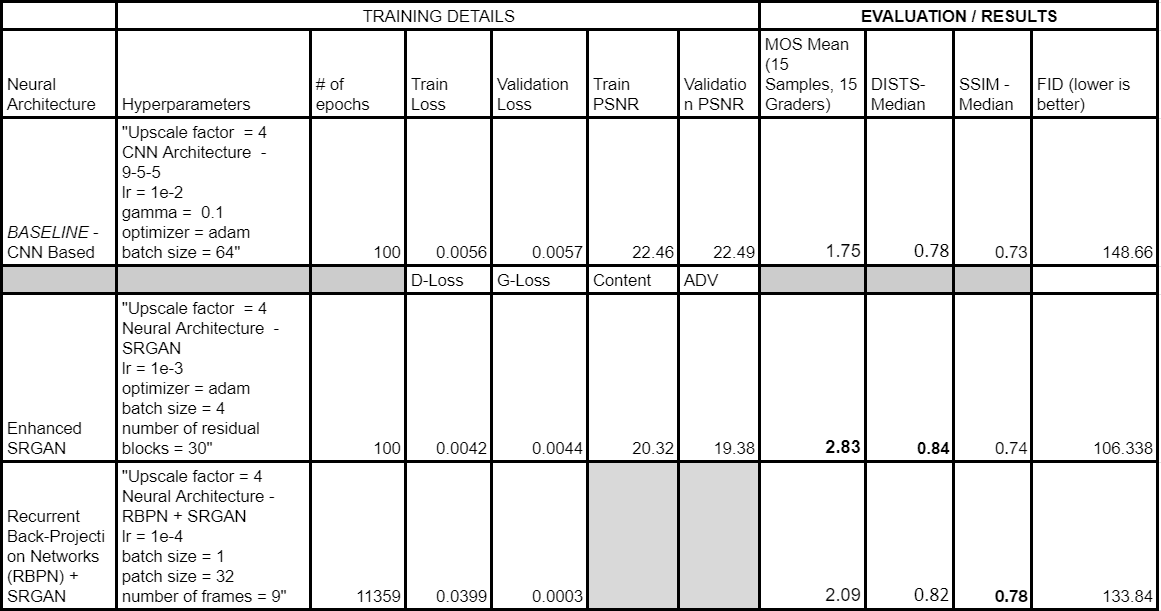}
\caption{Training and Scoring Hyper Parameters and Evaluation Metrics}
\end{center}
\end{figure}

\begin{figure}[!htbp]
\begin{center}
\includegraphics[scale=0.4]{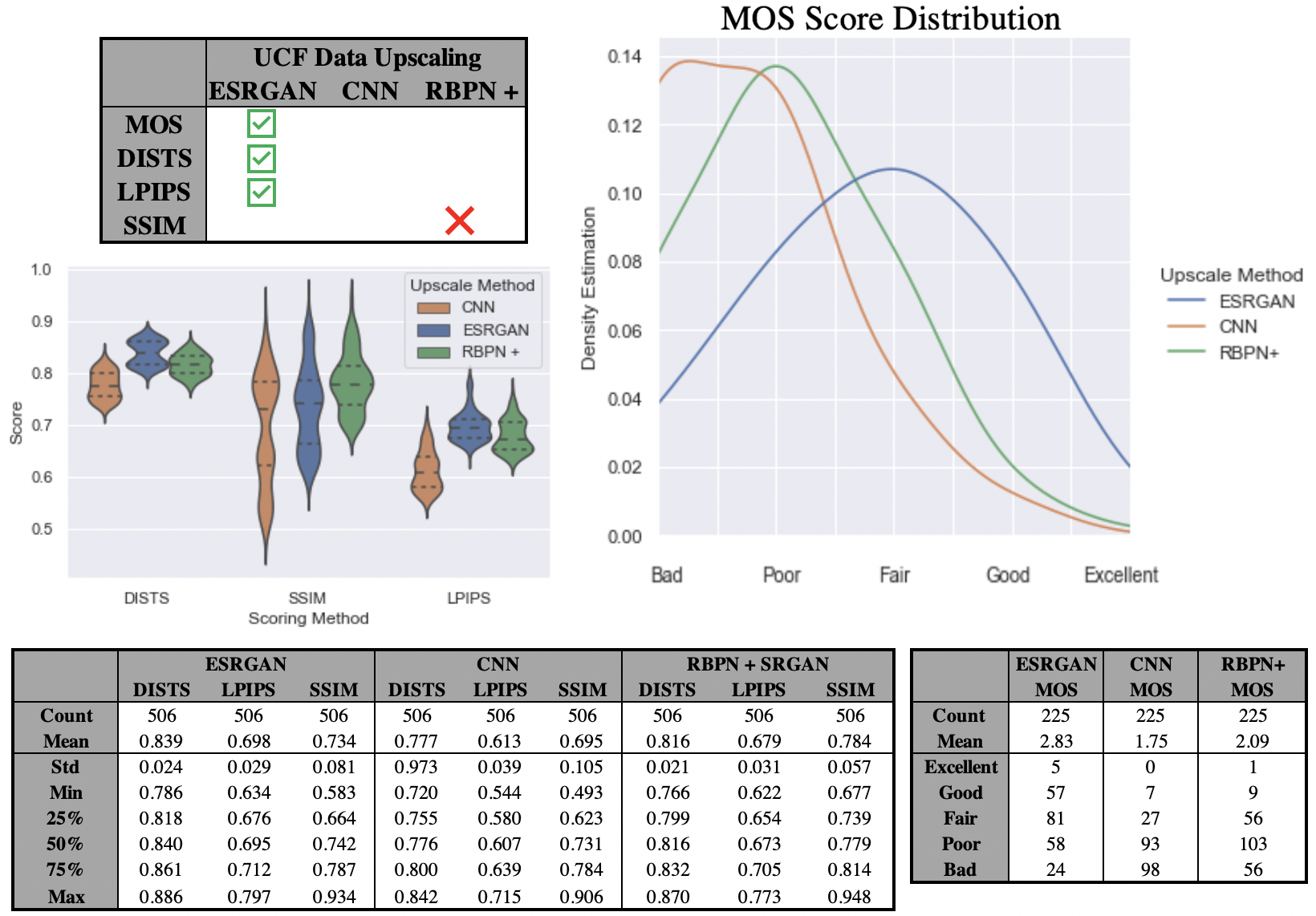}
\caption{Scoring Distributions on UCF Dataset}
\end{center}
\end{figure}

\begin{figure}[!htbp]
\begin{center}
\includegraphics[scale=0.5]{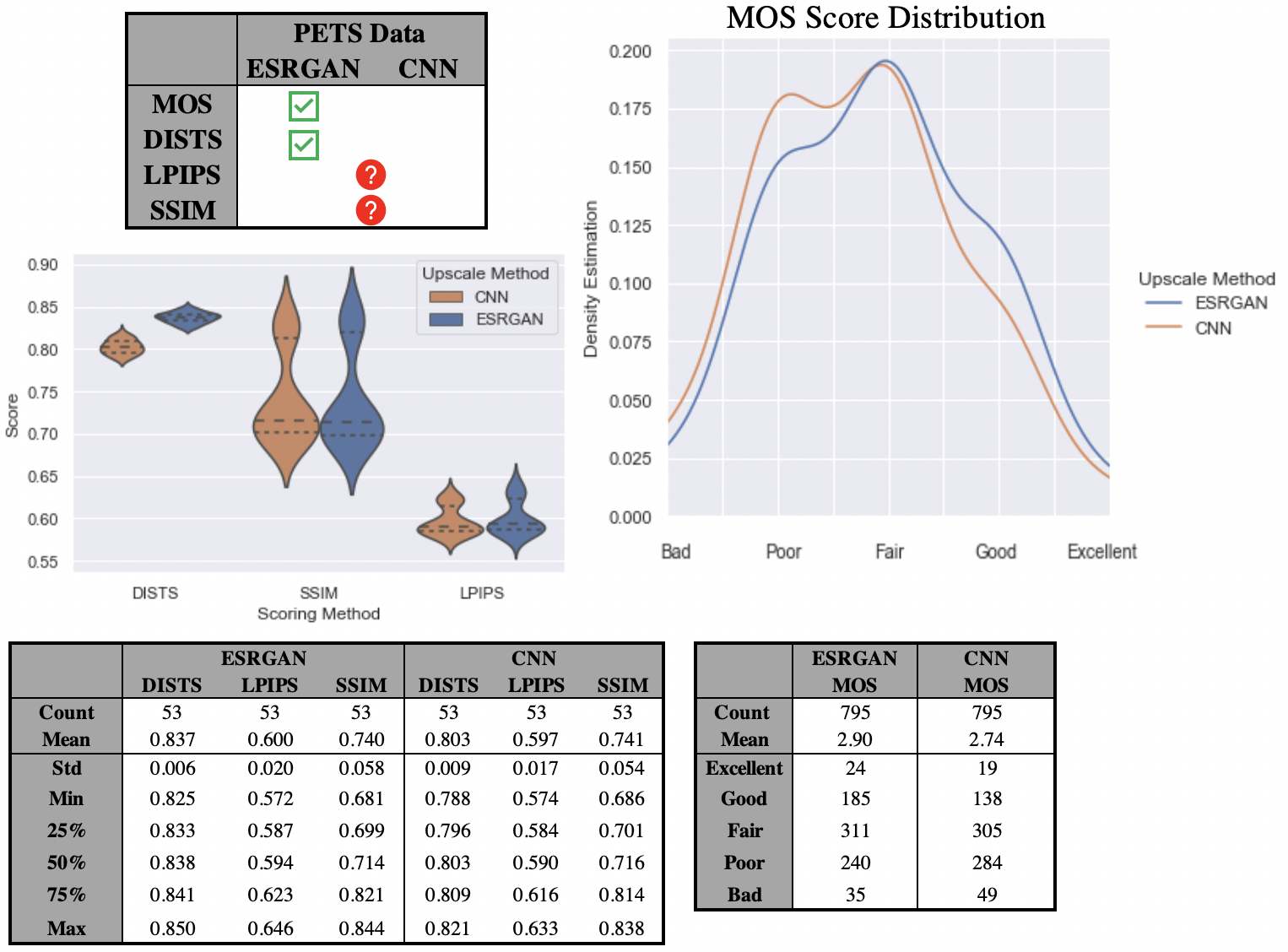}
\caption{Scoring Distributions on PETs dataset}
\end{center}
\end{figure}

To establish ground truth, we started by evaluating our datasets with MOS. In order to compensate for human biases and differences in environment, we randomly selected 15 frames from the UCF dataset which we held constant across the up-sampling methods. For each of these 15 frames we sourced 15 separate individuals via Mechanical Turk to provide a MOS score for that frame, in isolation of any other images, for a total of 225 respondents. To confirm out finding on the PETS dataset, we chose to gather the opinions of 15 different people on all 53 images, for a total of 795 respondents. Throughout both datasets, the ESRGAN was found to have higher MOS scores with statistical significance at an alpha of 0.001. 

For the UCF dataset, we tabulated the three objective IQA scoring metrics for each of the up-sampling methods. We found that both of the deep network scoring methods correctly predicted that the ESRGAN up-sampling method had the highest MOS scores; while SSIM incorrectly predicted the RBPN+ method had the best quality. The results are shown in figure 10.

To confirm our findings, we up-sampled the PETS dataset with the same up-scaling factor using both the ESRGAN and CNN methods and again scored the images using all three objective IQA metrics. This dataset had a more similar distribution of MOS scores between up-sampling methods, and was a good differentiator between the various IQA metrics. Only the DISTS metric was able to differentiate that the ESRGAN had higher quality at an alpha of 0.001. The other two objective IQA metrics could not reject the null hypothesis and were unclear on which up-sampling method would have higher MOS scores. The results are shown in figure 11. 

Additional as you can see in figure 9, the FID\cite{FID_NIPS2017_8a1d6947} score is consistent with both our MOS and DISTS scores, the FID score is lowest for our Enhanced SRGAN implementation. In case of FID score lower the score, closer it is to the original image in terms of quality\cite{FID_NIPS2017_8a1d6947}. This can also be seen in the figure 12. An important point to note here is that there are several drawbacks\cite{DBLP:journals/corr/abs-1802-03446} of FID\cite{FID_NIPS2017_8a1d6947} scores to which DISTS\cite{DBLP:journals/corr/abs-2004-07728} is more robust such as:
\begin{enumerate}
    \item FID\cite{FID_NIPS2017_8a1d6947} assumes that features are of Gaussian distribution which is often not guaranteed.
    \item Unlike IS\cite{simonyan2015deep} however, it is able to detect intra-class mode dropping\cite{DBLP:journals/corr/abs-1802-03446}, i.e. a model that generates only one image per class can score a high IS but will have a bad FID\cite{FID_NIPS2017_8a1d6947}.
    \item FID\cite{FID_NIPS2017_8a1d6947} worsens as various types of artifacts are added to images/\cite{DBLP:journals/corr/abs-1802-03446}
    \item Another drawback is since FID\cite{FID_NIPS2017_8a1d6947} uses inception features to calculate the score, the features may not accurately represent real world samples. 
\end{enumerate}

\begin{figure}
\begin{center}
\includegraphics[scale=0.3]{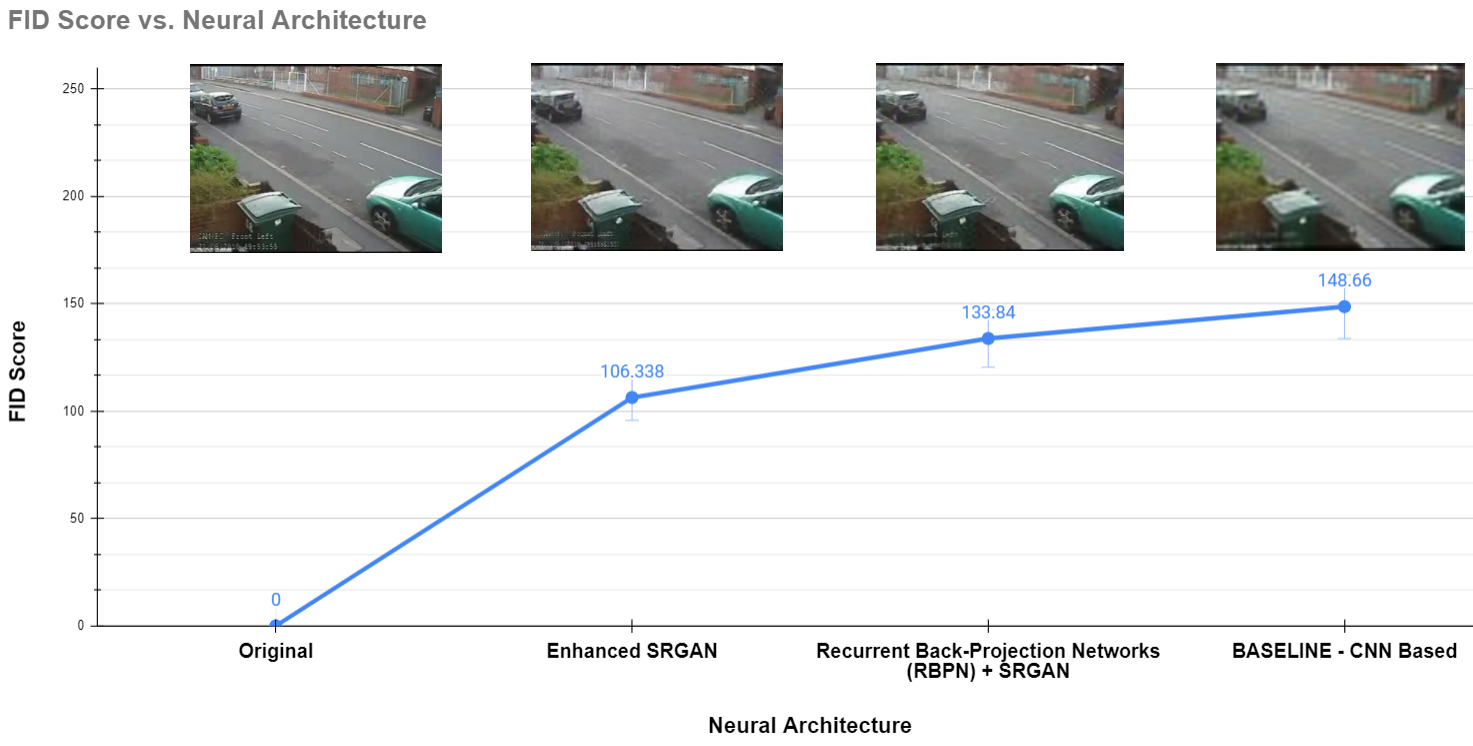}
\caption{Neural architectures vs FID Scores (lower is closer to original)}
\end{center}
\end{figure}

\section{Conclusion}
In conclusion we have bought forth two important observations and contributions. Firstly we show that Deep Image Structure and Texture Similarity (DISTS) correlates well with human quality judgments (Mean Opinion Scores - MOS\cite{MOS}) across two independent surveillance datasets(UCF\cite{sultani2019realworld} and PETS 2007\cite{pets_2007}) used to up-sample low resolution surveillance images and video using state of the art neural architectures. We have also experimentally shown that DISTS is a stronger metric to evaluate the up-sampling quality of surveillance videos and that DISTS is superior to both SSIM[27] and LPIPS[29] when trying to predict MOS[21] scores as seen in figures 10 and 11. 

Secondly, we evaluate three approaches for video up-sampling of surveillance videos on the UCF Mini \cite{sultani2019realworld} and PETS 2007\cite{pets_2007} datasets and we observe that the Enhanced SRGAN based approach produce the highest quality and resolution for the 4x up-sampled images and videos across the three approaches we evaluated as determined by MOS and again seen in figures 10 and 11. Another important observation is the scoring latency of these approaches; the Sub-pixel CNN Based Video Up-sampling performs 4x up-scaling at the lower latency, almost 10x faster than the GAN based approaches that we have implemented. This makes the Sub-pixel CNN Based Video Up-sampling a good fit for more real time surveillance scenarios with a trade-off of up-scaling quality, whereas the GAN based approach produce higher quality up-scaling with high latency, making them a good fit for offline up-sampling scenarios.

\section{Future Work}
Through this work, we strive to provide a novel evaluation and adaptation of state of the art neural architectures to up-sample low resolution surveillance images and videos as well as highlight DISTS as a stronger IQA evaluation metric for the same. Building on this work we plan to use our best performing model to up-sample the UCF Anomaly Videos \cite{sultani2019realworld} dataset to provide the research community a higher quality surveillance dataset to experiment on. In addition, we would like to investigate if iSeeBetter may be improved by using a DISTS inspired loss function vice the combined four loss functions that it is currently using. We also plan to build upon our experiments to apply them in the domain of surveillance scene and subject restoration using object detection and up-sampling using our models. 

\nocite{*}
\bibliographystyle{plain}
\bibliography{main}

\end{document}